\documentclass{article}

    \PassOptionsToPackage{square,sort,comma,numbers}{natbib}

    \usepackage[preprint]{neurips_2020}

\usepackage[utf8]{inputenc} 
\usepackage[T1]{fontenc}    
\usepackage{hyperref}       
\usepackage{url}            
\usepackage{booktabs}       
\usepackage{amsfonts}       
\usepackage{nicefrac}       
\usepackage{microtype}      
\usepackage{natbib}
\usepackage{graphicx}
\usepackage{amsfonts}
\usepackage{subfigure}
\usepackage{todonotes}
\usepackage{amsmath}
\usepackage{caption}
\usepackage{authblk}

\title{End-to-end Sinkhorn Autoencoder\\with Noise Generator}

\author[1]{Kamil Deja}
\author[1]{Jan Dubiński}
\author[1]{Piotr Nowak}
\author[2]{Sandro Wenzel}
\author[1,3]{Tomasz Trzciński}

\affil[1]{Warsaw University of Technology, pl. Politechniki 1 Warsaw, Poland}
\affil[2]{CERN, 1211 Geneva 23, Switzerland}
\affil[3]{Tooploox}
\affil[ ]{\{k.deja, t.trzcinski\}@ii.pw.edu.pl, jan.dubinski.stud@pw.edu.pl, \{piotr.wladyslaw.nowak,~sandro.wenzel\}@cern.ch}

\begin{document}

\maketitle

\begin{abstract}
    In this work, we propose a novel end-to-end sinkhorn autoencoder with noise generator for efficient data collection simulation. Simulating processes that aim at collecting experimental data is crucial for multiple real-life applications, including nuclear medicine, astronomy and high energy physics. Contemporary methods, such as Monte Carlo algorithms, provide high-fidelity results at a price of high computational cost. Multiple attempts are taken to reduce this burden, e.g. using generative approaches based on \textit{Generative Adversarial Networks} or \textit{Variational Autoencoders}. Although such methods are much faster, they are often unstable in training and do not allow sampling from an entire data distribution. To address these shortcomings, we introduce a novel method dubbed \textit{end-to-end Sinkhorn Autoencoder}, that leverages sinkhorn algorithm to explicitly align distribution of encoded real data examples and generated noise. More precisely, we extend autoencoder architecture by adding a deterministic neural network trained to map noise from a known distribution onto autoencoder latent space representing data distribution. We optimise the entire model jointly. Our method outperforms competing approaches on a challenging dataset of simulation data from Zero Degree Calorimeters of ALICE experiment in LHC. as well as standard benchmarks, such as MNIST and CelebA.
    
    
    
\end{abstract}

\section{Introduction}

Multiple real-life applications rely heavily on detailed simulations of ongoing processes, from atomic structures in nuclear medicine (e.g. tomography)~\cite{strulab2003gate} or genetcis \cite{incerti2018geant4}, to astrophysics~\cite{zoglauer2006megalib}. 
This is also true for the Large Hadron Collider (LHC)~\cite{Evans:2008zzb} -- one of the biggest scientific programmes currently being carried out worldwide. In the LHC two beams of particles are accelerated to the ultra-relativistic energies and brought to collide. In such environment, high energy density leads to the appearance of very rare phenomena. 
To understand these processes, physicists compare recorded data with accurate theoretical models simulations. 
Currently employed simulation techniques use complex Monte Carlo processing in order to compute all possible interactions between particles and matter. Such an approach produces accurate results at the expense of high computational cost. 

Therefore multiple attempts are taken to speed up this processing, including those that leverage state of the art generative models~\cite{paganini2017calogan, sofia18, deja2018generative} such as Generative Adversarial Networks~\cite{goodfellow2014generative} (GANs) or Variational Autoencoders~\cite{kingma2013auto} (VAE).

While above methods are much faster than standard simulations, they suffer from the limitations of generative models.
In principle 
training of Generative Adversarial Networks is often unstable and may result in limited quantitative properties~\cite{arora2017generalization,arora2017gans}. On the other hand, Variational autoencoders converge much better, however they also provide worse alignment between original data distribution and generated one. 
Because of the maximum likelihood approximation and sampling from enforced distribution, generative models based on autoencoders have both visual and statistical problems with reconstructing the original data distribution.

To stabilise the training of GANs, in~\cite{arjovsky2017wasserstein} authors propose to substitute KL divergence with Wasserstein distance.
Since it was proven to be more reliable a new model dubbed Wasserstein Autoencoder (WAE)~\cite{tolstikhin2017wasserstein} was proposed, where the same metric is used to regularise distribution of data in autoencoder's latent space. In WAE authors employed two different methods to calculate wasserstein distance -- MMD and adversarial critic. While original processing of WAE provides high quality results, a new autoencoder based architectures such as Sliced Wasserstein Autoencoder~\cite{kolouri2018sliced} or Sinkhorn Autoencoder~\cite{patrini2018sinkhorn} introduce faster non-adversarial wasserstein distance approximations.

   

Thanks to the autoencoder based processing, denoted techniques provide stable training. However, they require significant regularisation on the original autoencoder's latent space, which enables sampling from parametrised distribution. This regularisation enforces encoding in a particular hyperspace 
which often leads to limited representation capabilities. 
In particular, commonly used 
Normal distribution 
hinders linear separability of different components of complex distribution~\cite{glorot2011deep}. It also does not allow sparse representation obtained via relu activation~\cite{tonolini2019variational}.


\begin{figure}
\begin{minipage}[c]{0.65\linewidth}
 \centering
  \includegraphics[width=.99\linewidth]{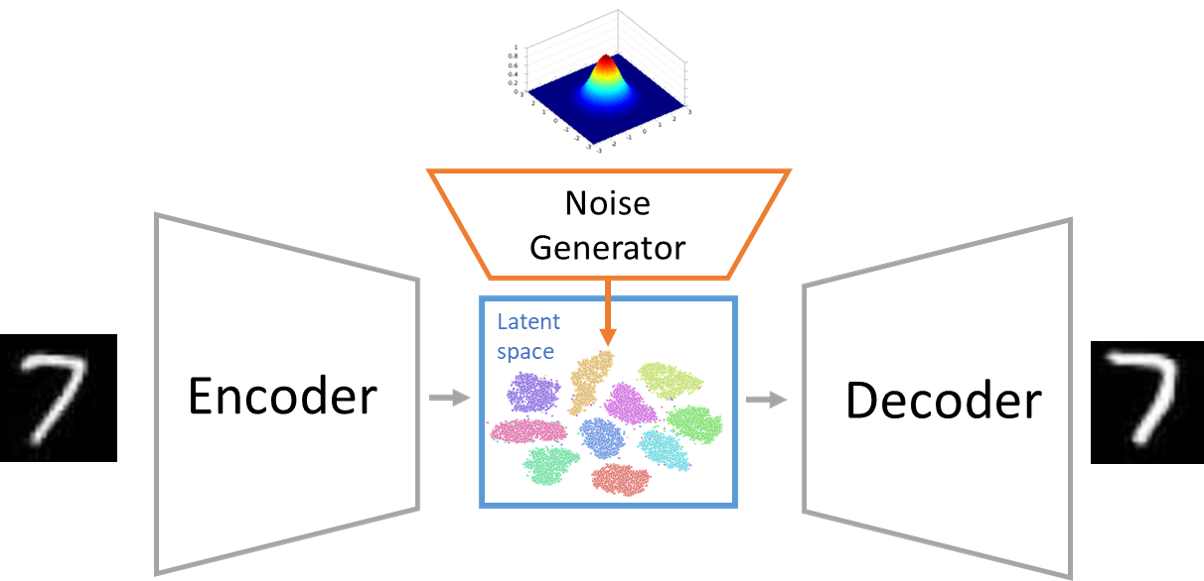}
\end{minipage}
\begin{minipage}[c]{0.34\linewidth}
 \centering
  \includegraphics[width=.9\linewidth]{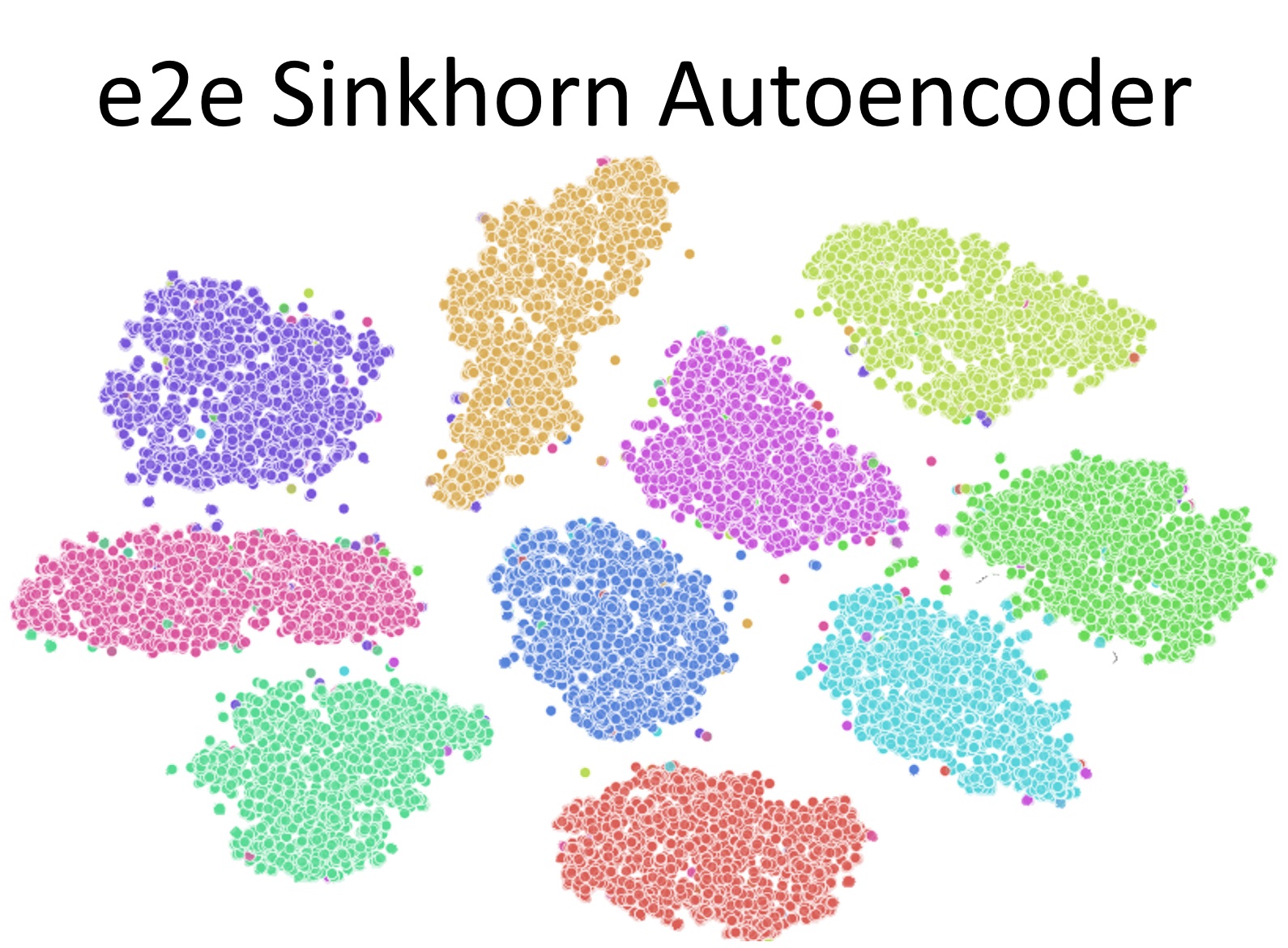}
  \includegraphics[width=.9\linewidth]{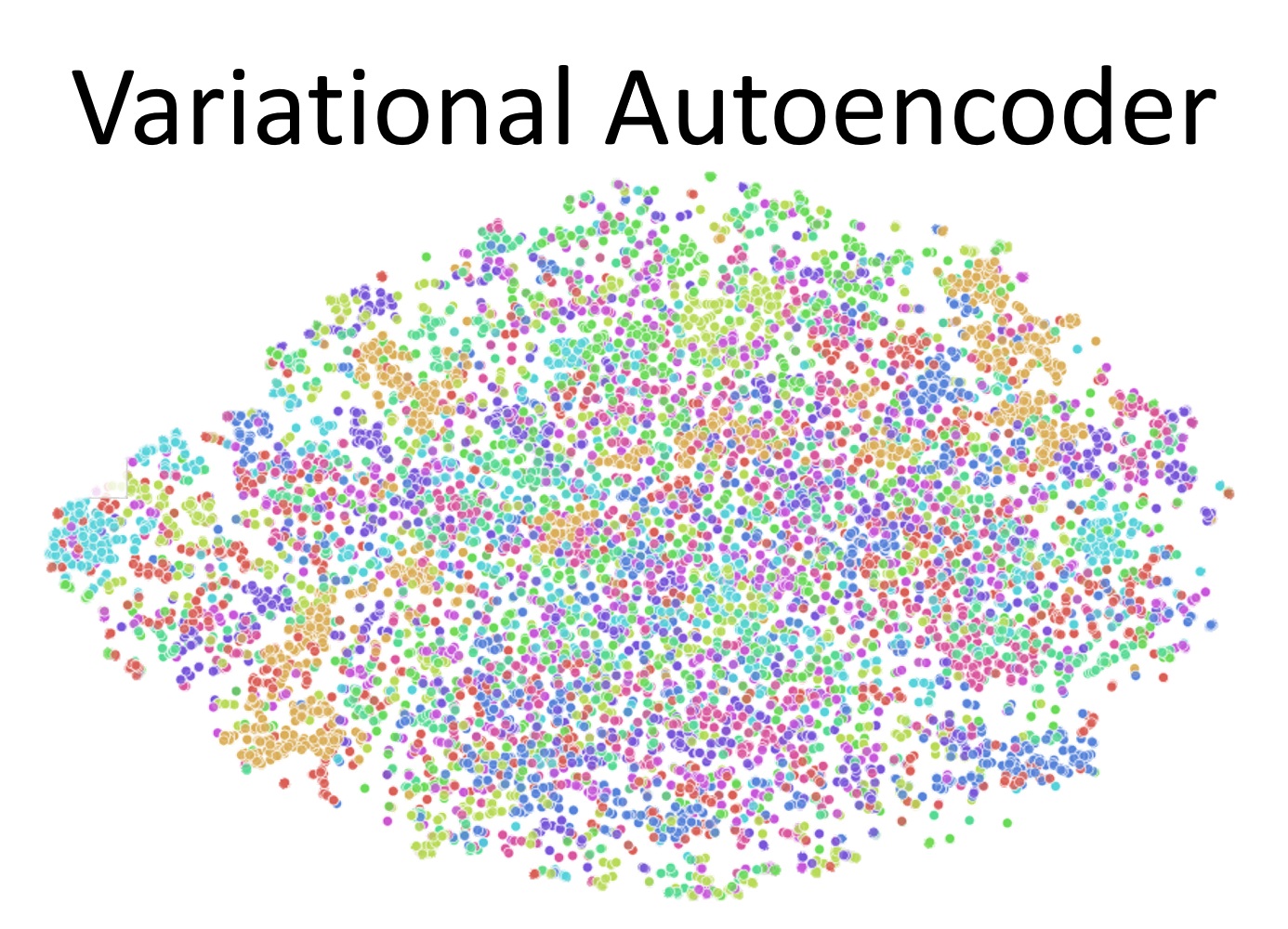}
\end{minipage}
\caption{\label{fig:schema}Schematic visualisation of end-to-end sinkhorn autoencoder processing (left). TSNE visualisation of latent space for mnist dataset (right). Our conditional e2e sinkhorn autoencoder (top) and conditional VAE (bottom). Our model does not restrict latent space to the normal distribution, therefore classes may be even linearly separable.}
\end{figure}

To address these shortcomings we propose a new generative model build on top of the sinkhorn autoencoder~\cite{patrini2018sinkhorn}.
Instead of restricting  autoencoder to encode examples on the parametrised distribution, we approximate it 
with explicit \textit{noise generator} implemented through additional deterministic neural network, as presented in Fig.~\ref{fig:schema}. We input noise from a known distribution (\textit{e.g.} Normal) to this network and encode it to follow distribution of real data in autoencoder's latent space. Although such approach allows us to generate new data samples from a parametrised distribution, thanks to additional neural network we do not regularise our encoders latent space with such constrains
In our approach we only require that autoencoder's latent space can be approximated with a non-linear mapping, from a known distribution. This is a much softer assumption than the one present in current solutions.
To our knowledge our end to end sinkhorn autoencoder is the first generative autoencoder without explicit constraint on the latent space.


Problem of constrained autoencoder latent space is even more evident in conditional generative models, as it hinders separation based on a priori information.
Currently proposed conditional generative models such as conditional VAE~\cite{sohn2015learning} include additional a priori parameters to the encoder and decoder. At the same time condVAE regularises latent space to follow normal distribution. Therefore model learns to encode information related to classes only in encoder and decoder, while in latent space all of the examples are shuffled into a single manifold as presented in Fig.~\ref{fig:schema}. This behaviour limits  classes separation, since they have to be learned in decoder from one common continuous distribution.



In this work we introduce conditional version of our solution. Contrary to prior methods, we do not input conditional parameters into encoder and decoder. 
We allow autoencoder to encode different classes in different areas of latent space, while we match them with conditional noise generator. Such an approach, is more suitable for different (e.g. imbalanced) conditional classes. It allows to encode data into more natural, disentangled representation with clear classes separation as depicted in figure Fig.~\ref{fig:schema}.%

We evaluate the quality of our standard and conditional end-to-end sinkhorn autoencoder with commonly used benchmark datasets, such as MNIST~\cite{lecun1998gradient} and CelebA~\cite{liu2015deep}, and achieve state-of-the-art results. 
To show generalisation of our solution we then apply the processing to the problem of fast simulation of particle showers in High Energy Physics.
We show that our method allows to generate high-quality caloremeter responses and significant coverage in original data distribution. The superiority of our model is even more pronounced on this dataset.


The main contributions of this work are:
\begin{itemize}
    \item Introduction of a new non-adversarial end-to-end generative model with explicit noise generator.
    \item New conditional generative model based on an autoencoder architecture which on the contrary to the currently employed models leave the structure of autoencoder's latent space intact.
\end{itemize}

The remainder of this work is organised as follows. 
In Sec.~\ref{sec:related} we describe related works in the field of autoencoding generative modelling and fast simulations for HEP. Sec.~\ref{sec:solution} outlies our end-to-end sinkhorn autoencoder method followed by its conditional version. We conclude this work in Sec.~\ref{sec:experminets}, with experiments on MNIST, CelebA and HEP datasets and description of potential further studies. 

\section{\label{sec:related} Related work}


Traditional autoencoder architecture is a method commonly used for dimensionality reduction~\cite{wang2016auto}, representation learning~\cite{vincent2010stacked} or anomalies detection~\cite{sakurada2014anomaly} (also in HEP~\cite{pol2019anomaly}). Its direct application in generative modelling is hard because of the natural tendency to encode examples into complex latent space. Without regularisation such encoding often results in non-overlapping distribution with discontinuities as in Fig.~\ref{fig:schema}. 
While it allows for accurate 
reconstructions it makes sampling from a latent space nearly impossible.

Therefore different generative models based on autoencoders are trained to regularise encoder so that the latent space is continuous and follows a parameterised distribution. Authors of the first in the field Variational Autoencoder~\cite{kingma2013auto} propose regularisation on latent space based on KL divergence $D_{KL}(P_X,P_Z)$, where $P_X$ is the original encoded data distribution, and $P_Z$ is the prior distribution. In practice, for $P_Z$ authors propose Gaussian generators. 
Wasserstein Autoencoders~\cite{tolstikhin2017wasserstein} changes the regularisation objective from KL divergence to wasserstein distance. In this work authors introduce two possible methods for application of wasserstain distance on autoencoder's latent space. First one is based on the \textit{maximum mean discrepancy}~\cite{gretton12a} (MMD) technique, while second one similarly to~\cite{arjovsky2017wasserstein} uses neural network as a critic. 
However, following~\cite{kingma2013auto}, authors also use Gaussian generators as priors.
In our end-to-end sinkhorn autoencoder we benefit from the processing of those standard models. However, we use faster and more stable approximations of wasserstein distance and we do not regularise autoencoder's latent space to the prior distribution. 

Sliced Wasserstein Autoencoder~\cite{kolouri2018sliced} proposes to substitute MMD with approximation obtained by cumulative distribution of one-dimensional distances. This solution is much simpler, but as reported in~\cite{patrini2018sinkhorn} it results in lower diversity of generated results. 
Therefore, we build our model on top of Sinkhorn Autoencoder~\cite{patrini2018sinkhorn}. In~\cite{patrini2018sinkhorn}
authors use yet another wasserstein distance approximation based on the sinkhorn algorithm~\cite{cuturi2013sinkhorn}.
As in previous attempts VAE or WAE, sinkhorn autoencoder maps original data encodings to the selected prior $P_Z$. In particular, in~\cite{patrini2018sinkhorn} authors experiment with normal and hypersphere prior. In this work we benefit from sinkhorn autoencoder's analysis to investigate further, proposed by the authors sweet-spot for generative atuoencoding research. However, instead of regularising latent space with a prior we converge to it with a trainable noise generator.




We introduce our model together with its application for the problem of calorimeters response simulations. 
Majority of current works in this field focus on GAN architectures, like a very first CaloGAN~\cite{paganini2017calogan} or~\cite{sofia18}. In those works authors adapt conditional DCGAN~\cite{radford2015unsupervised} architecture. 
We tackle the same problem from different perspective using autoencoding generative model. Therefore, in Sec.~\ref{sec:experminets} we compare our results to the DCGAN model which is a basis of above solution.

\section{\label{sec:solution} Sinkhorn autoencoder with noise generator}

In this section we describe our new end-to-end sinkhorn autoencoder generative model. At first we depict its general topology, then we examine three parts of final model optimisation objective: reconstruction loss, sinkhorn loss on latent space and regularisations. Finally we introduce conditional version of our model.

In Sinkhorn Autoencoder work~\cite{patrini2018sinkhorn} authors introduce a generative model framework in which they employ sinkhorn algorithm to match autoencoder's latent space with known distribution.
In our work we leverage this analysis and present an extended version of this method with trainable prior approximator dubbed \textit{noise generator}. We implement it through deterministic neural network.
With such approach we can use gradient obtained from sinkorn loss to simultaneously train encoder and noise generator to converge into similar distribution on the latent space.

\begin{figure}[htb]
	\centering
	\includegraphics[width=0.85\linewidth]{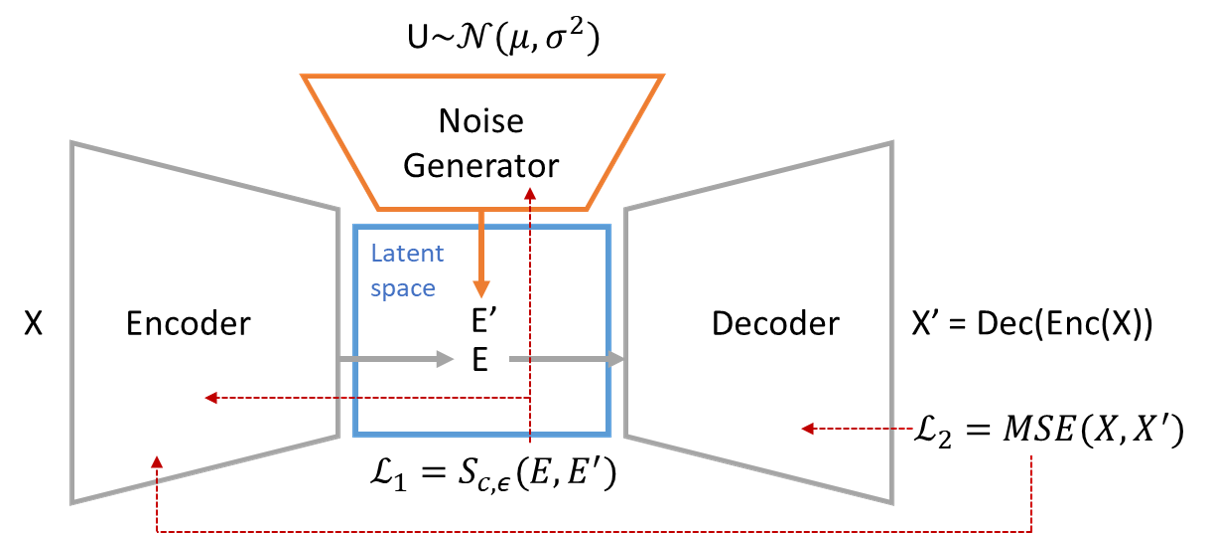}
	\caption{Architecture of the sinkhorn autoencoder with neural network as an explicit noise generator. Red arrows indicate the gradient flow. Reconstruction Loss $L_2$ is backpropagated through decoder and encoder, while sinkhorn loss $L_1$ is propagated in two directions to encoder and noise generator. Encoder network is optimised with a sum of $L_1$ and $L_2$ losses.}
	\label{fig:arch}
	\vspace{-0.5cm}
\end{figure}

In Fig.~\ref{fig:arch} we demostrate the general layout of our solution. It consists of three neural networks -- \textit{encoder}, \textit{decoder} and \textit{noise generator}. As presented in Fig.~\ref{fig:arch}, loss of our model composes of two main terms - $L_1$ - sinkhorn loss on latent space and $L_2$ reconstruction loss of the autoencoder. 
Additionally, to prevent overfitting and promote diversity, we employ regularisations on both model weights and autoencoder's latent space. 

\subsection{Reconstruction loss}
The core of our network is based on standard autoencoder, hence it follows original autoencoder training procedure. Encoder is trained to map the original data $X$ into the latent space $E$, while decoder is optimised to reconstruct original data examples $X'$. In this part we experiment with two different losses.
Standard Mean Squared Error loss $MSE(X,X')=(X-X')^2$ is a simple choice, but as denoted in~\cite{bojanowski2017optimizing} it may lead to the blurry images. Therefore, following~\cite{bojanowski2017optimizing} for training on certain datasets such as CelebA, we also employ Laplacian pyramid $Lap_1$ loss presented in below:

\begin{equation}
    Lap_1(x,x') = \sum_{j} 2^{2j}|L^j(x) - L^j(x')|_1
    \label{eq:lap}
\end{equation}
where $L^j(x)$ is the j-th level of the Laplacian pyramid representation of x~\cite{ling2006diffusion}.

Similarly to ~\cite{bojanowski2017optimizing} as a final reconstruction objective we use a weighted mean of the standard mean squared error and the $Lap_1$ loss.

\begin{equation}
    L_{recon}(X,X') = \alpha Lap_1(X,X')+ MSE(X,X') 
    \label{eq:recon_loss}
\end{equation}
where $\alpha$ is a scaling parameter. 

\subsection{Sinkhorn loss}

Similarly to~\cite{patrini2018sinkhorn}, we leverage sinkhorn  algorithm~\cite{cuturi2013sinkhorn} to regularise our latent space. However, on the contrary to~\cite{patrini2018sinkhorn} we use gradient obtained from this loss to train both our \textit{encoder} and \textit{noise generator} to encode/generate embeddings with the same distribution in the latent space. 
This computation proceeds as follows. First, we encode the batch of original images $X$ to obtain their encoded representation $E$. At the same time we process a random vector $U$ sampled from a known distribution (e.g. $\mathcal{N}(0,1)$) through the noise generator. It creates noise representation $E'$ in the same latent space as for encoded images.

Then we compute the distance between real data and noise embeddings. 
Following WGAN or WAE architectures we could approximate this with additional neural network, but to simplify the solution we opt for entropy regularisation of wasserstein distance implemented with sinkhorn algorithm. 

For this purpose we follow ~\cite{genevay2017learning,genevay2018sample} to define the entropy regularized Optimal Transport cost with $\epsilon \geq 0$ as:
\begin{equation}
    \tilde S_{c,\epsilon}(P_X,P_Y) = \inf_{\Gamma \in \Pi(P_X,P_Y)} \mathbb{E}_{(X,Y) \sim \Gamma}[c(X,Y)] + \epsilon \cdot KL(\Gamma,P_X \otimes P_Y).
\end{equation}

As suggested in~\cite{patrini2018sinkhorn} we remove the entropic bias of the above approximation with three passes of sinkhorn algorithm, as presented below:

\begin{equation}
    S_{c,\epsilon}(P_X,P_Y) = \tilde S{c,\epsilon}(P_X,P_Y) - \frac{1}{2}(\tilde S{c,\epsilon}(P_X,P_X) + \tilde S{c,\epsilon}(P_Y,P_Y))
    \label{eq:sinkhorn_final}
\end{equation}

With the above equation we calculate the loss value for two representations of encoded images $E$ and generated noise $E'$ in a batch-wise manner as $S_{c,\epsilon}(E,E')$. For the wasserstein cost function $c$ we use standard 2-Wasserstein distance with euclidean norm $c(x,y)=\frac{1}{2}||x-y||^2_2$. As indicated in ~\cite{genevay2018sample} $S_{c,\epsilon}$ deviates from the original Wasserstein distance by approximately $O(\epsilon log(1/\epsilon)$, hence we keep our $\epsilon$ small to avoid the influence on network's convergence.
In practice we use the efficient implementation of sinkhorn algorithm with GPU acceleration from GeomLoss package~\cite{feydy2019interpolating}. 

\subsection{End-to-end sinkhorn autoencoder objective}

To improve the diversity of generated images we include additional regularisation on the autoencoders latent space. For this purpose we adapt diversity regularisation proposed in~\cite{ayinde2019regularizing}
. In this work authors compute a similarity matrix $SIM_{C}$ to asses the diversity in the neural network's weights according to the cosine similarity between its different layers.


We adapt this technique in our model and measure the similarity between all of encoded real data examples from the batch. Then, following~\cite{ayinde2019regularizing} we compute the regularisation as a sum of these similarities as presented in the equation~\ref{eq:div_reg}:

\begin{equation}
    R_{s}({\mathbf{y}})= p \sum_{i=1}^{bs} \sum_{j=1, j \neq i}^{bs} m_{i, j}\left({SIM}_{C}\left({\mathbf{y}_{i}}, \mathbf{y}^T_j\right)\right)^{2}
    \label{eq:div_reg}
\end{equation}

where $p$ is a scaling factor $bs$ is batch size and $m$ is a binary mask variable which drops pairs below threshold $\tau$.
\begin{equation}
m_{i, j}=\left\{\begin{array}{ll}
1, & \left|{SIM}_{C}\left(\mathbf{y}_{i}, \mathbf{y}^T_j\right)\right| \geq \tau \\
0, &  otherwise
\end{array}\right.
\end{equation}

Additionally we also experiment with different regularisations on autoencoders weights. In our experiments we observed better convergance with $L_2$ regularisation on last layer of our encoder.

Below we outline the joint loss function of our autoencoder as a sum of four elements: reconstruction loss, sinkhorn loss between generated noise and original encoded images and additional regularisations on network's latent space and weights values $\theta$.

\begin{equation}
    \begin{split}
        L_{sum}(X) = & \alpha L_{recon}(X,dec(enc(X)))
        \\&+ \beta     S{c,\epsilon}(enc(X),gen(X'\sim\mathcal{N}(0,1)) \\
        &+ \delta R_s(enc(X)) +\gamma reg(\theta) 
    \end{split}
    \label{eq:autoenc_final}
\end{equation}

\subsection{Conditional Sinkhorn objective}

While the goal for most applications of generative modelling is to generate more examples from a given distribution, for certain tasks we have to include additional \textit{a priori} information about simulated data. This is also the case for HEP, where we want to simulate possible responses of calorimeter for a given particle.
For the purpose of conditional images generation we propose a simple adjustment to the standard version of our processing presented in the previous section.

Firstly, for a given batch of samples $X$ with corresponding conditioning variables $Q$ we propose to include conditional values in the \textit{noise generator} as a separate input to the neural network. Thanks to this, we change our \textit{noise generator} to conditional form, since it will encode random noise $E'|Q$ with respect to conditional values $Q$. 
Secondly, we train the \textit{noise generator} and \textit{encoder} to map examples with similar conditional values near to each other in the latent space. 
For that purpose, 
we first encode all of the examples $X$ into the their representation in latent space $E$. Then, for each example $e \in E$ we concatenate its encoding with corresponding conditional values $q \in Q$. We perform the same operation for noise encodings $E'$ and their conditional parameters $Q$ obtained from original training data. Finally we pass concatenated vectors through the sinkhorn algorithm, calculating the loss value.

\begin{equation}
    S_{c,\epsilon}(enc(X)^\frown Q,gen(X'\sim\mathcal{N}(0,1),Q)^\frown Q)
\end{equation}


Thanks to this approach we add additional cost to the original Wassserstein objective. 
Specifically this cost for the $Wasserstein_2$ metric is the euclidean  norm distance between different conditional values. However, depending on the nature of the a priori information $Q$, and potential real cost of generating samples from the distribution of wrong conditions, it might be beneficial to change it or scale original values.

With this approach on the contrary to the other conditional generative models such as conditional VAE or WAE, 
our solution leaves the original autoencoder's latent space intact. We do not enforce it to encode different classes into the general consistent distribution
Thanks to the fact that conditional parameters are included in noise generator we can observe that autoencoder distribute different classes in separate areas of its latent space. We compare both of exemplar latent spaces for MNIST dataset in figure Fig.~\ref{fig:schema}.

\section{\label{sec:experminets}Experiments}

In this section, we evaluate performance of our end-to-end sinkhorn autoencoder model in reference to other generative solutions. Primarily, we compare results of our experiments to other autoencoder based generative approaches. For that purpose we use two standard benchmarks: the MNIST dataset of handwritten digits~\cite{lecun1998gradient} and CelebA dataset of celebrity face images~\cite{liu2015deep}, cropped to 64x64 pixels. We also evaluate different generative solutions including adversarial example of Deep Convolutional GAN~\cite{radford2015unsupervised} on a challenging dataset of calorimeters response simulations which we call HEP. It consists of 117817 44x44 pixels images together with 9 a priori attributes on colliding particle (mass, momenta, charge, energy). 

To assess models quality we use \textit{Fréchet inception distance} (FID) introduced in~\cite{heusel2017gans}. As proposed in~\cite{binkowski2018demystifying} for MNIST dataset we change the original Inception neural network to LeNet based convolutional classifier. 
While FID is criticised~\cite{shmelkov2018good} for approximating distributions with Gaussians, we also introduce a new measure to monitor diversity of generated examples. After computing response from LeNet network we compare obtained distributions with wasserstein distance approximation implemented with sinkhorn algorithm. We refer to this measurement as \textit{Sinkhorn}.

For HEP dataset we benefit from the simulation nature of the data to asses its quality 
on the basis of the physical properties of generated output. According to the specification~\cite{dellacasa1999alice}, we compute a sum of specific pixels of generated image dubbed \textit{channels}, which represent simulated collision.   
Then we compare the distance between generated results and ground truth values from real simulations. This allows us to measure how accurate our model is in terms of data generation with respect to conditional values. Additionally we measure quality 
of reconstructing full spectrum of channel values. For this purpose we measure Wasserstein distance between original and generated channels distribution.

In our experiments we follow networks architectures proposed in Wasserstein Autoencoder~\cite{tolstikhin2017wasserstein}. Therefore for CelebA dataset we use convolutional deep neural network with 4 convolutional/deconvolutional layers for both encoder and decoder with 5x5 filters. Additionally we use batch normalisation~\cite{ioffe2015batch} after each convolutional layer. For our noise generator we use simple fully connected network with 3 hidden layers and ReLU activations. We optimise our networks with Adam~\cite{kingma2014adam} in batches of 1000 examples.\footnote{Code for experiments is available at \url{https://github.com/ulr_hidden_for_review_process}}

\begin{table}
  \caption{Results comparison for conditional generative models on MNIST and HEP datasets}
  \label{results_mnist}
  \centering
  \begin{tabular}{l|ll|ll}
    \toprule
    &\multicolumn{2}{c|}{MNIST} &\multicolumn{2}{c}{HEP}               \\      
    \cmidrule(r){1-5}
    Model     & FID     & Sinkhorn& MAE &Wasserstein \\
    \midrule
    cond VAE & 6.61 &30.13 & 23.13 ($\pm 65.53$) & 14.92  \\
    cond WAE (MMD) & 34.73 & 30.44  &  43.54 ($\pm$55.23)& 34.46\\ 
    cond e2eSAE (ours)  & \textbf{4.11 }& \textbf{24.92 } & \textbf{13.50 ($\pm$29.82)} & \textbf{7.91} \\
    \cmidrule(r){1-5}
    cond DCGAN     & 0.93  & 22.23 & 68.27 ($\pm 180.45$) & 6.95 \\
    original data & 0.33  & 0 & 6.59 & 2.89 \\
    \bottomrule
  \end{tabular}
\end{table}

\begin{figure}[bt]
	\centering
	\includegraphics[width=0.65\linewidth]{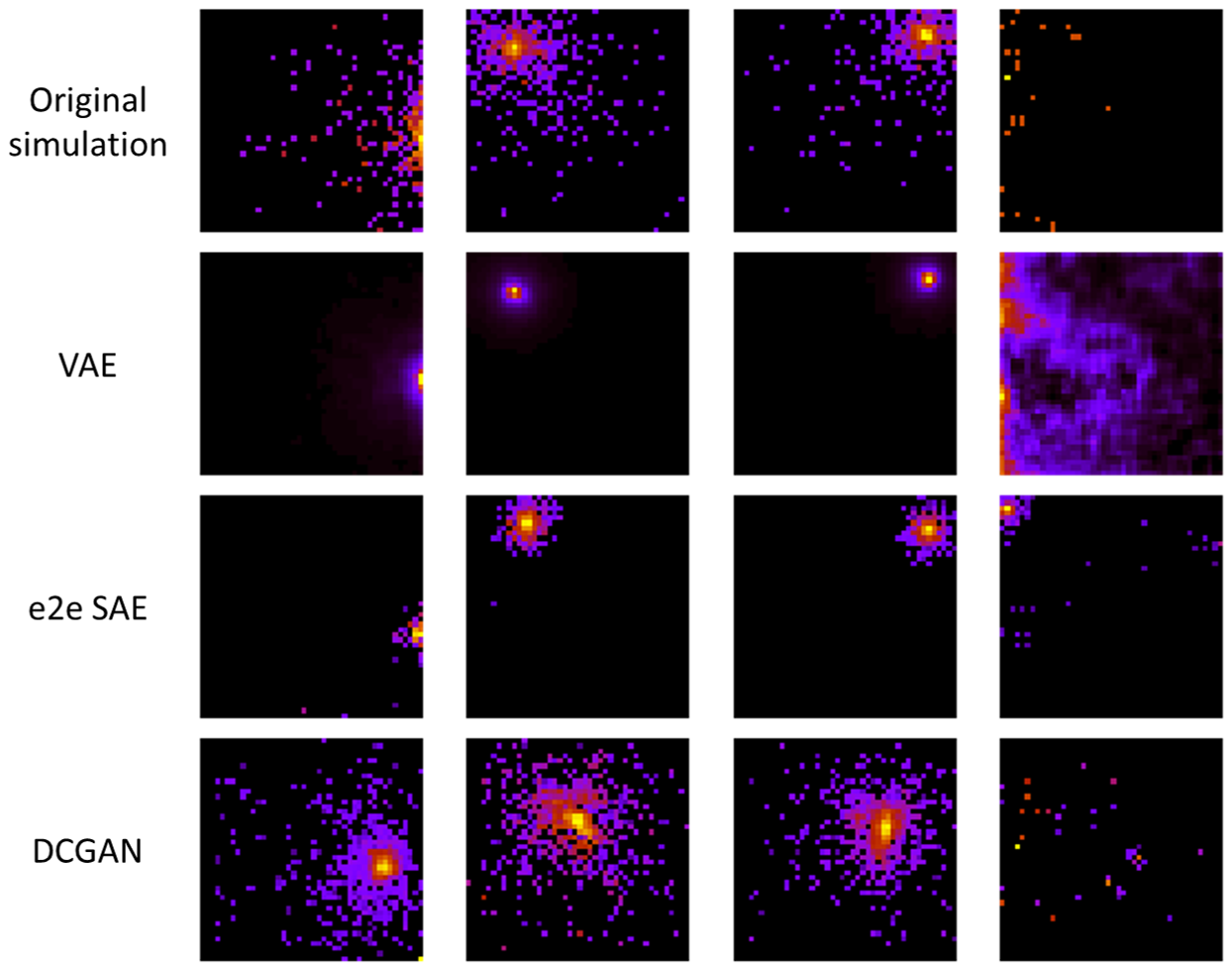}
	\caption{Examples of calorimeters response simulations with different methods. Although results from GAN are visually sound with collisions, model was not able to properly capture relations from conditional values. Our solution did not reproduce all of residual values, it outperformed other methods in terms of accuracy for the most significant centre of collision.}
	\label{fig:zdc_examples}
\end{figure}

As presented in in Tab.~\ref{results_mnist}, our conditional model outperforms other non-adversarial solutions on both MNIST and HEP datasets. As shown in Fig.~\ref{fig:zdc_examples}, HEP dataset remains challenging for all generative models. For VAE we can see the outcome of regularisation with normal distribution. Although our solution does not provide results as visually attractive as conditional DCGAN, it better captures relations between conditional parameters. We believe this is because of the specific nature of simulations which allows unlimited values for pixels.

\begin{figure}[htb]
		\centering
		\subfigure[]{\label{fig:dcgan_tsne}\includegraphics[width=0.325\linewidth]{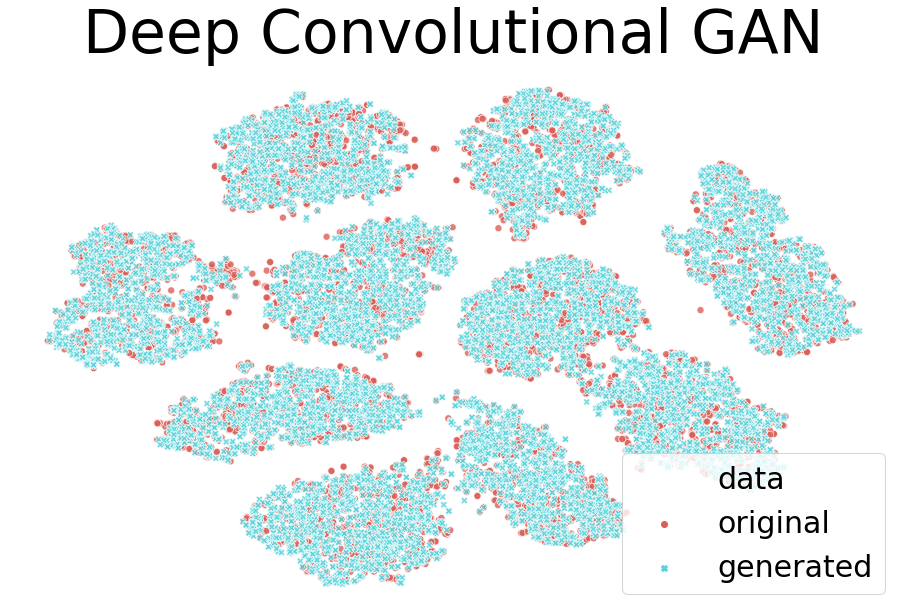}}
		\subfigure[]{\label{fig:vae_tsne} \includegraphics[width=0.325\linewidth]{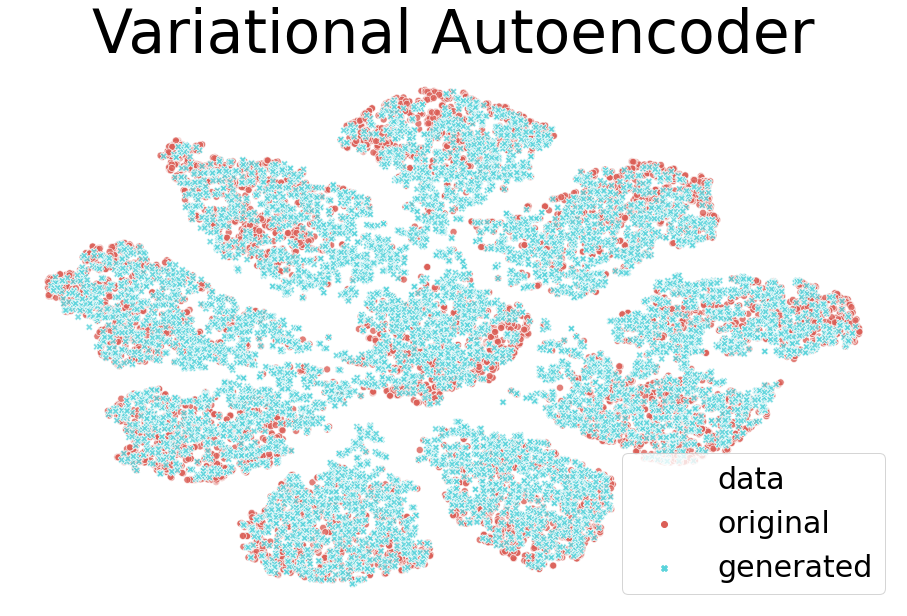}}
		\subfigure[]{\label{fig:sae_tsne}\includegraphics[width=0.325\linewidth]{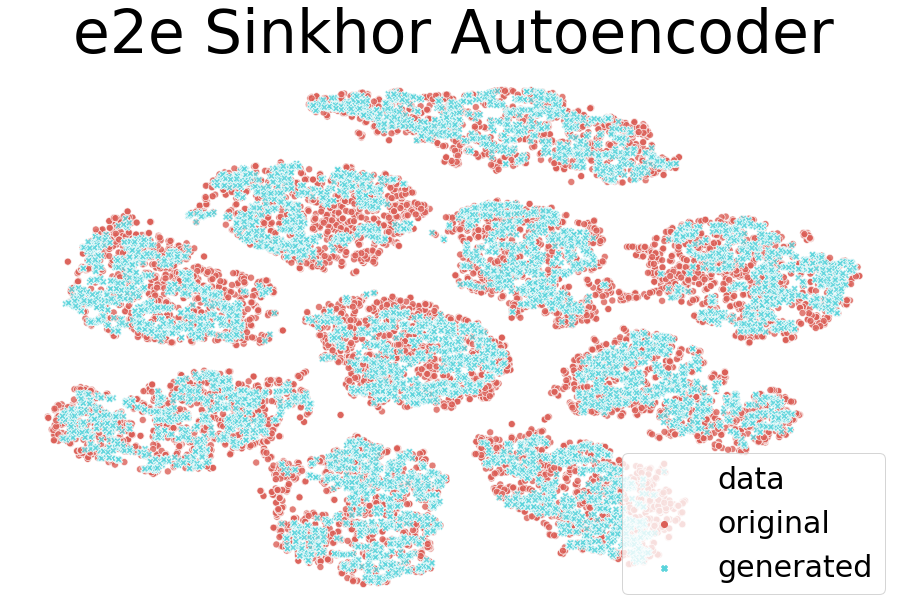}}
		\caption{\label{fig:tsne}TSNE visualisation of generated examples (blue) and original mnist data (red), processed through LeNet network. DCGAN reproduces the whole data distribution well, while VAE additionally produces images from outside of real data distribution. Our solution (right) generates only examples within true data distribution but does not properly reproduce the whole variety.}
\end{figure}

For MNIST, we also visually analyse coverage of data distribution for evaluated methods. As presented in TSNE visualisation of LeNet penultimate layer activations in Fig.~\ref{fig:tsne}, our method has better coverage of original data then other autoencoder based approaches. As depicted in Fig.~\ref{fig:sae_tsne}, our model does not produce examples outside of original data distribution. On the other hand, results from the well trained adversarial model (Fig.~\ref{fig:dcgan_tsne}) better overlaps with full data distribution of MNIST dataset.

\begin{figure}[hbt]
\begin{minipage}{0.6\linewidth}
	\centering
	\includegraphics[width=.8\linewidth]{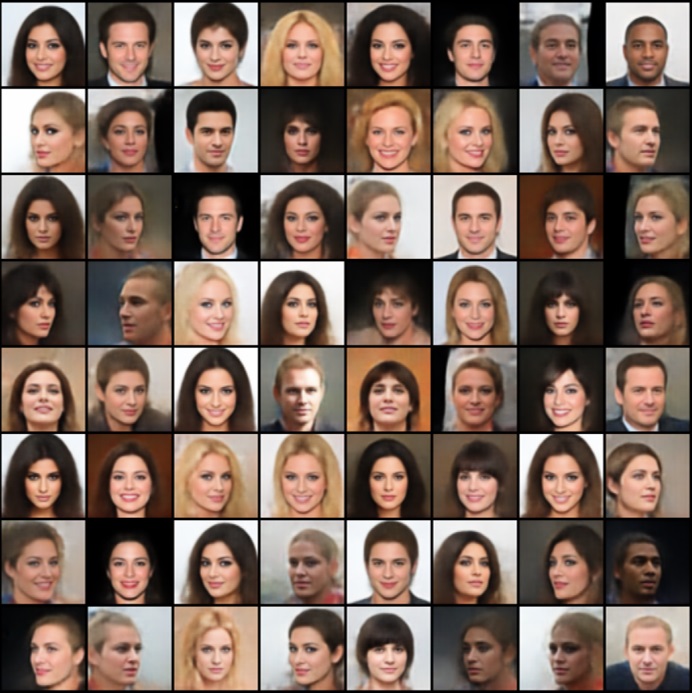}
	\captionof{figure}{Samples of generated images from model trained on CelebA dataset. Our model is capable of generating diverse, high quality images without blurred effect.}
	\label{fig:celeba_examples}
\end{minipage}
\begin{minipage}{0.39\linewidth}
  \caption{Results comparison on CelebA dataset. For competitive solutions we include the best of reported result.}
  \label{tab:celeba_results}
  \centering
  \begin{tabular}{ll}
    \toprule
    \multicolumn{2}{c}{CelebA}             \\      
    \cmidrule(r){1-2}
    Model     & FID \\
    \midrule
    VAE & 55\\
    WAE (MMD)& 55 \\ 
    SWAE & 64\\
    SAE ($H$) & 56\\
    e2e SAE (ours)& \textbf{54.5}  \\
    \bottomrule
  \end{tabular}
\end{minipage}
\end{figure}

On CelebA dataset as demonstrated in table~\ref{tab:celeba_results} our method outperforms other competitive autoencoder based solutions. As displayed in figure Fig.~\ref{fig:celeba_examples} our end-to-end sinkhorn autoencoder generates visually sharp images with high variance. 

\section{Conclusions}

In this work, we introduced a new generative model based on autoencoder architecture. Contrary to contemporary solutions, our end-to-end sinkhorn autoencoder does not enforce encoding on any paremetrised distribution. In order to learn distribution of standard autoencoder we converge to it with additional deterministic neural network, which we train together with autoencoder. We show that our solution outperforms other comparable approaches on benchmark datasets as well as challenging dateset of calorimeter response simulations. We postulate that processing proposed in this work may also be used with other metrics between probability distribution.






\section*{Acknowledgements}

Authors acknowledge the support from the Polish National Science Centre grants no. UMO-2018/31/N/ST6/02374 and UMO-2016/21/D/ST6/01946. 
The GPUs used in this work were funded by the grant of the Dean of the Faculty of Electronics and Information Technology at Warsaw University of Technology (project II/2017/GD/1). 

\section*{Broader Impact}

Large Hadron Collider is stepping into the new era. During long shutdown till 2021 a lot of components change in order to improve performance of the collider.
\textit{E.g.} After the next start of operation in 2021, ALICE should be able to register 3 orders of magnitude more collision than its current capabilities  (from 50 Hz to 50 kHz)~\cite{abelev2014upgrade,o2}.

At the same time currently employed Monte Carlo based simulations require high computational resources, hence they already occupy 50-70\% of all CERN computing power worldwide. Therefore major changes are needed in terms of true data simulation techniques in order to follow LHC technical upgrades. 

Development of fast simulation tools is crucial the for whole branch of High Energy Physics. However, exactly the same techniques may also be used in the nuclear medicine treatment. Furthermore, other disciplines may also take advantage of the fundamental research conducted in HEP experiments.



\bibliographystyle{plain}
\bibliography{references}

\newpage
\section*{Appendix}

In this appendix we attach high quality generations for CelebA dataset  as well as additional examples from MNIST dataset which are not included in the original article.

\begin{figure}[hbt]
	\centering
	\subfigure[]{\includegraphics[height=5cm]{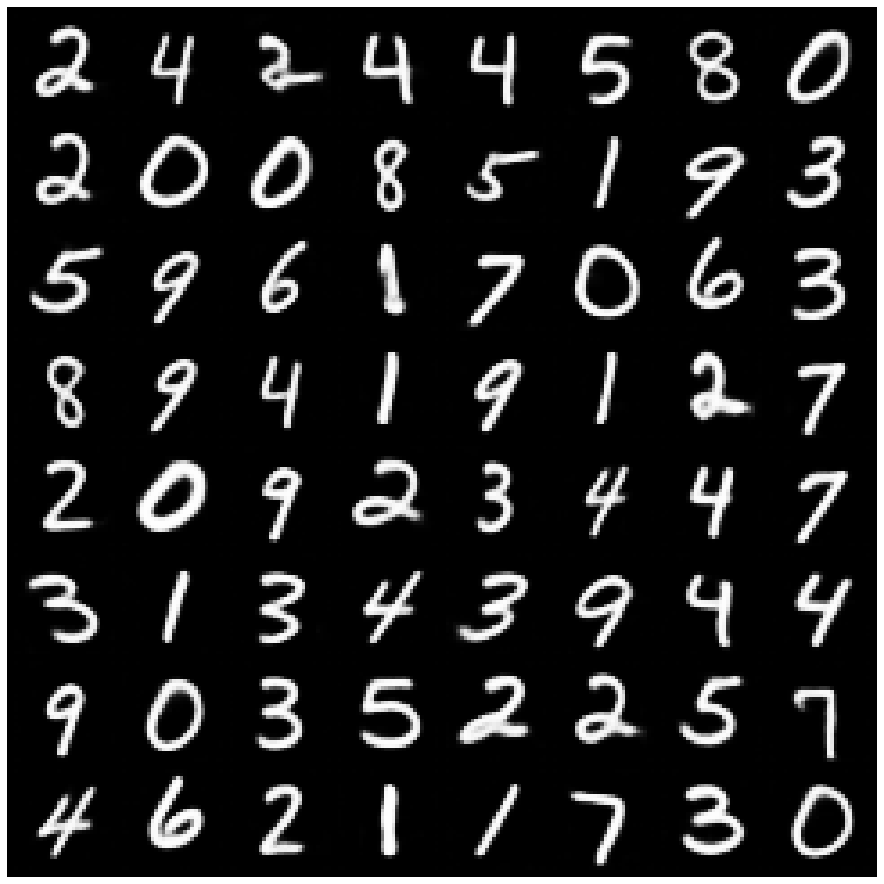}}
	\subfigure[]{\includegraphics[height=5cm]{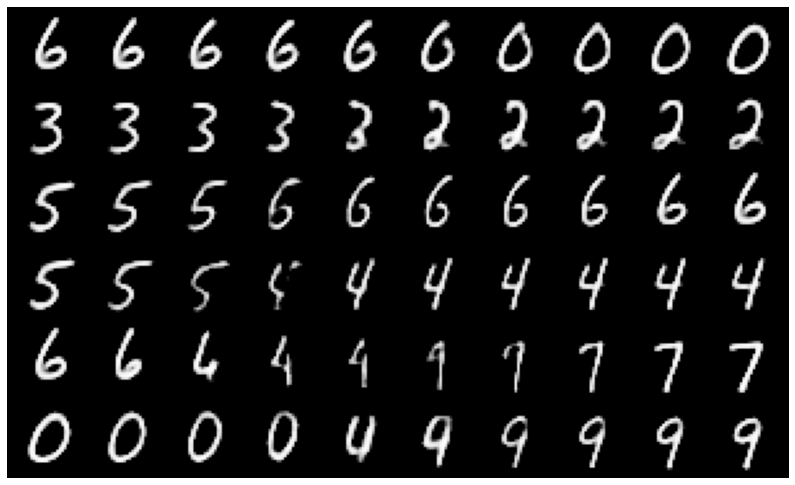}}
	\caption{Samples of generated numbers from the model trained on MNIST dataset(left) and interpolation between different classes(right). Our conditional end-to-end sinkhorn autoencoder creates various images. During interpolation it does not create examples from outside of original classes even for invalid conditional parameters.}
	\label{fig:interpolation}
\end{figure}

\begin{figure}[hbt]
	\centering
	\includegraphics[width=.8\linewidth]{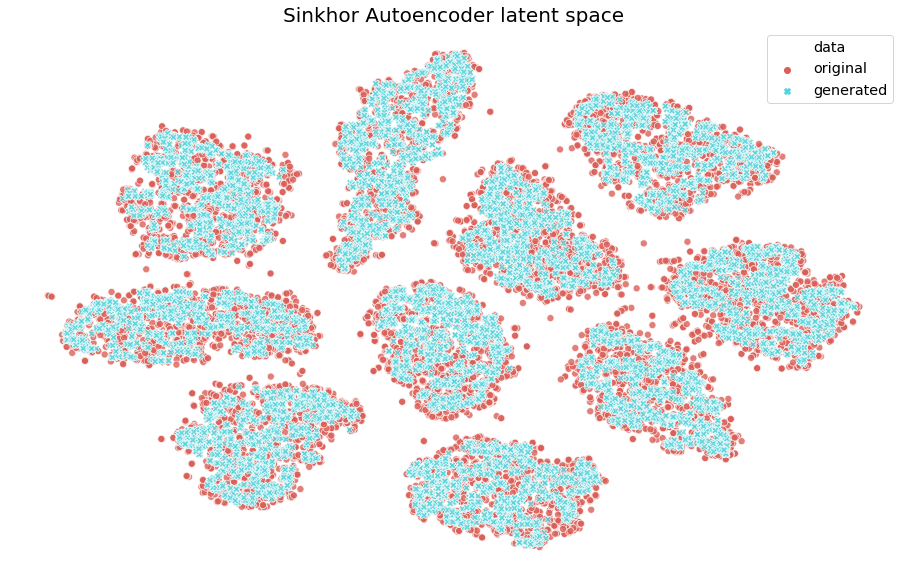}
	\captionof{figure}{TSNE visualisation of our end to end sinkhorn autoencoder's latent space trained on MNIST dataset -- original values (red) and noise generated with properly trained noise generator (blue). As visible, deterministic neural network trained with sinkhorn loss was able to properly learn mapping between normal noise and real data examples representation on latent space.}
	\label{fig:latent_generated}
\end{figure}

\begin{figure}[hbt]
	\centering
	\includegraphics[width=.9\linewidth]{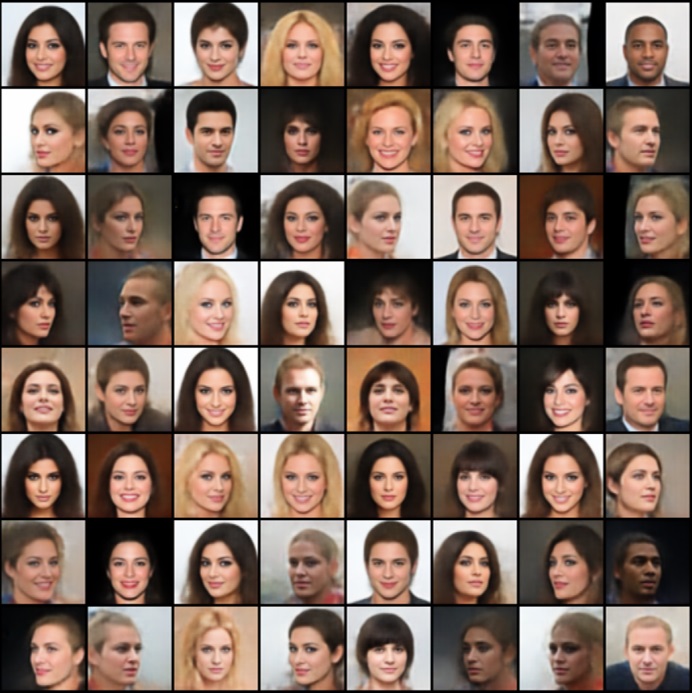}
	\captionof{figure}{Samples of generated images from model trained on CelebA dataset. Our model is capable of generating diverse, high quality images without blurred effect.}
	\label{fig:celeba_examples_big}
\end{figure}

\end{document}